\documentclass{article} 
\usepackage{iclr2016_workshop, times} 
\usepackage{hyperref}
\usepackage{url}
\usepackage{graphicx}
\usepackage{tikz}
\usepackage{amsmath}
\usepackage[outercaption]{sidecap}
\usepackage[T1]{fontenc}

\title{Blending {LSTM}s into CNNs}

\author{Krzysztof J. Geras$\mathbf{^{1}}$, Abdel-rahman Mohamed$\mathbf{^{2}}$, Rich Caruana$\mathbf{^{2}}$, Gregor Urban$\mathbf{^{3}}$, \\ \textbf{Shengjie Wang$\mathbf{^{4}}$, \"{O}zlem Aslan$\mathbf{^{5}}$, Matthai Philipose$\mathbf{^{2}}$, Matthew Richardson$\mathbf{^{2}}$ \& Charles Sutton$\mathbf{^{1}}$}\\
$^{1}$University of Edinburgh\\
$^{2}$Microsoft Research\\
$^{3}$UC Irvine\\
$^{4}$University of Washington\\
$^{5}$University of Alberta
}

\newcommand{\eg}{e.g.\ }

\newcommand*\idx[2][]
{ 
\def\next{#1}%
\ifx\empty\next
  (#2)
\else
  (#1, #2)
\fi
}
\newcommand*\elt[3][]
{ 
\def\next{#1}%
\ifx\empty\next
  #2\idx{#3}
\else
  #1\idx{#2,#3}
\fi
}
\newcommand*\pd[3][]
{ 
\def\next{#1}%
\ifx\empty\next
  \frac{\partial#2}{\partial #3}
\else
  \frac{{\partial^{#1} #2}}{\partial#3^{#1}}
\fi
}

\newcommand{\hfor}{\overrightarrow{h}}
\newcommand{\hback}{\overleftarrow{h}}
\newcommand{\igate}{i}
\newcommand{\fgate}{f}
\newcommand{\ogate}{o}
\newcommand{\state}{c}
\newcommand{\hiddenfn}{\mathcal{H}}

\newcommand{\wtmat}[2]{W_{#1 #2}}
\newcommand{\ihwts}{\wtmat{x}{h}}
\newcommand{\hhwts}{\wtmat{h}{h}}

\newcommand{\bias}[1]{b_{#1}}
\newcommand{\hbias}{\bias{h}}
\newcommand{\obias}{\bias{y}}
\newcommand{\seq}[1]{\boldsymbol #1}

\newcommand{\invble}{x}

\newcommand{\inseq}{\seq{\invble}}

\begin{document}

\maketitle

\begin{abstract}
We consider whether deep convolutional networks (CNNs) can 
represent decision functions with similar accuracy as recurrent networks such as LSTMs.
First, we show that a deep CNN with an architecture inspired by the models recently introduced in image recognition can yield better accuracy than previous convolutional and LSTM networks on the standard 309h Switchboard automatic speech recognition task.
Then we show that even more accurate CNNs can be trained under the guidance of LSTMs
using a variant of model compression, which we call \emph{model blending}
because the teacher and student models are similar in complexity but different in inductive bias.
Blending further improves the accuracy of our CNN, yielding a computationally efficient model of accuracy higher than any of the other individual models. Examining the effect of
``dark knowledge'' in this model compression task, we find that less than 1\% of the
highest probability labels are needed for accurate model compression.
\end{abstract}

\section{Introduction}

There is evidence that feedforward neural networks trained with current training algorithms use their large capacity inefficiently \citep{optimal_brain, predicting_parameters, big_waste, do_deep_nets, distilling_knowledge, deep_compression}. Although this excess capacity may be necessary for accurate learning and generalization at training time, the function once learned often can be represented much more compactly. As deep neural net models become larger, their accuracy often increases, but the difficulty of deploying them also rises. Methods such as model compression sometimes allow the accurate functions learned by large, complex models to be compressed into smaller models that are computationally more efficient at runtime.  

There are a number of different kinds of deep neural networks such as deep fully-connected neural networks (DNNs), convolutional neural networks (CNNs) and recurrent neural networks (RNNs). Different domains typically benefit from deep models of different types. For example, CNNs usually yield highest accuracy in domains such as image recognition where the input forms a regular 1, 2, or 3-D image plane with structure that is partially invariant to shifts in position or scale. On the other hand, recurrent network models such as LSTMs appear to be better suited to applications such as speech recognition or language modeling where inputs form sequences of varying lengths with short and long-range interactions of different scales.

The differences between these deep learning architectures raise interesting questions about what is learnable by different kinds of deep models, and when it is possible for a deep model of one kind to represent and learn the function learned by a different kind of deep model. The success of model compression on feedforward networks raises the question of whether other neural network architectures that embody different inductive biases can also be compressed. For example, can a classification function learnt by an LSTM be represented by a CNN with a wide enough window to span the important long-range interactions?

In this paper we demonstrate that, for a speech recognition task, it is possible to train very accurate CNN models, outperforming LSTMs. This is thanks to CNN architectures inspired by recent developments in computer vision \citep{very_deep_vision}, which were not previously considered for speech recognition. Moreover, the experiments suggest that LSTMs and CNNs learn different functions when trained on the same data. This difference creates an opportunity: by merging the functions learned by CNNs and LSTMs into a single model we can obtain better accuracy than either model class achieved independently. One way to perform this merge is through a variant of model compression that we call \emph{model blending} because the student and teacher models are of comparable size and have complementary
inductive biases. Although at this point model compression is a well established technique for models of different capacity, the question of whether compression is also effective for models of similar capacity has not been explored. For example, by blending an LSTM teacher with a CNN student, we are able to train a CNN that is more accurate because it benefits from what the LSTM learned and also computationally more efficient than an LSTM would be at runtime. This blending process is somewhat analogous to forming an ensemble of LSTMs and CNNs and then training a student CNN to mimic the ensemble, but in blending no explicit ensemble of LSTMs and CNNs need be formed. The blended model is \emph{6.8 times} more efficient at testing time than the analogous ensemble.
Blending further improves the accuracy of our convolutional model, yielding a computationally efficient model of accuracy higher than any of the other individual models. 
Examining the effect of ``dark knowledge'' \citep{distilling_knowledge} in this model compression task, we find that only 0.3\% of the
highest probability labels are needed during the model compression procedure. Intriguingly, we also find that model blending is even effective in the self-teacheing setting when the student and the teacher are of the same architecture. We show results of an experiment in which we use a CNN to teach another CNN of the same architecture. Such a student CNN is weaker than a CNN student of the LSTMs but still significantly stronger than a baseline trained only with the hard labels in the data set.

\section{Background}
\label{sec:background}

In model compression \citep{model_compression}, one model (a \emph{student}) is trained to mimic another model (a \emph{teacher}). Typically, the student model is small and the teacher is a larger, more powerful model, which has high accuracy but is computationally too expensive to use at test time. For classification, this mimicry can be performed in two ways. One way is to train the student model to match logits (i.e. the values $z_i$ in the output layer of the network, before applying the softmax to compute the output class probabilities $p_i = {e^{z_i}}/\sum_j{e^{z_j}}$) predicted by the teacher on the training data, penalising the difference between logits of the two models with a squared loss. Alternatively, compression can be done by training the student model to match class probabilities predicted by the teacher, by penalising cross-entropy between predictions $\mathbf{p}$ of the teacher and predictions $\mathbf{q}$ of the student, i.e. by minimising -$\sum_{i} p_i(\mathbf{x})\log q_i(\mathbf{x})$ averaged over training examples. We will refer to predictions made the teacher as \emph{soft labels}. In the context of deep neural networks, this approach to model compression is also known as \emph{knowledge distillation} \citep{distilling_knowledge}. Additionally, the training loss can also include the loss on the original 0-1 hard labels.

The main advantage of training the student using model compression is that a student trained with knowledge provided by the teacher gets a richer supervision signal than just the hard 0-1 labels in the training data, i.e. for each training example, it gets the information not only about the correct class but also about uncertainty, i.e., how similar the current training example is to those of other classes. Model compression can be viewed as a way to transfer inductive biases between models. For example, in the case of compressing deep models into shallow ones \citep{do_deep_nets, do_deep_convnets}, the student is benefiting from the hierarchical representation learned in the deep model, despite not being able to learn it on its own from hard labels.

While model compression can be applied to arbitrary classifiers producing probabilistic predictions, with the recent success of deep neural networks, work on model compression focused on compressing large deep neural networks or ensembles thereof into smaller ones, i.e., with less layers, less hidden units or less parameters. Pursuing that direction, \citet{do_deep_nets} showed that an ensemble of deep neural networks with few convolutional layers can be compressed into a single layer network as accurate as a deep one. In a complementary work, \citet{distilling_knowledge} focused on compressing ensembles of deep networks into deep networks of the same architecture. They also experimented with softening predictions of the teacher by dividing the logits by a constant greater than one called \emph{temperature}. Using the techniques developed in prior work and adding an extra mimic layer in the middle of the student network, \citet{fitnets} demonstrated that a moderately deep and wide convolutional network can be compressed into a deeper and narrower convolutional network with much fewer parameters than the teacher network while also increasing accuracy.

\subsection{Bidirectional LSTM}

One example of a very powerful neural network architecture yielding state-of-the-art performance on a range of tasks, yet expensive to run at test time, is the long short-term memory network (LSTM) \citep{lstm, bidirectional_LSTM, bidirectional_speech}, which is a type of recurrent neural network (RNN). The focus of this work is to use this model as a teacher for model compression. 

LSTMs exhibit superior performance not only in speech, but also in handwriting recognition and generation \citep{hand_recognition, hand_generation}, machine translation \citep{sequence_to_sequence} and parsing \citep{grammar}, thanks to their ability to learn longer-range interactions. For acoustic modeling though, the difference between a non-recurrent network and an LSTM using full-length sequences is two fold: the use of longer context while deciding on the current frame label, and the type of processing in each cell (the LSTM cell compared to a sigmoid or ReLU). The LSTM network used in our paper uses a fixed-size input sequence and only predicts the output for the middle item of the input sequence. In this respect, we follow the design proposed in the speech literature by \citet{baseline_LSTM}. We use acoustic models that use the same context window as a non-recurrent network (limited to about 0.5 s) while using the LSTM cells for processing each frame. The LSTM cells process frames in the same bidirectional manner that any other bidirectional LSTM would do, but they are limited by the size of the contextual window. Details of the LSTM used in this work can be found in the supplementary material. This style of modelling has important benefits. One motivation to use such an architecture is that acoustic modeling labels (i.e. target states) are local in nature with an average duration of about 450 ms. Therefore, the amount of information about the class label decays rapidly as we move away from the target. Long-term relations between labels, on the other hand, are handled using a language model (during testing) or a lattice of competing hypotheses in case of lattice training \citep{sequence_discriminative, lattice}. Another motivation to prefer models that utilise limited input windows is faster convergence due to the ability to randomise samples on the frame level rather than on the utterance level. A practical benefit of using a fixed-length window is that using bidirectional architectures becomes possible in real time setups when the delay in response cannot be long. 

\section{Vision-style CNNs for speech recognition}

Convolutional neural networks \citep{cnns} were considered for speech for many years \citep{conv_1998, convolutional_dbn}, though only recently have become very successful \citep{applying_convolutional, LVCSR, abdel_convolutional, LVCSR2}. These CNN architectures are quite different from those used in computer vision. They use only two or three convolutional layers with large filters followed by more fully connected layers. They also only use convolution or pooling over one dimension, either time or frequency. When looking at a spectrogram in \autoref{fig:example_utterance}, it is obvious that, like what we observe in vision, similar patterns re-occur both across different points in time and across different frequencies. Using convolution or pooling across only one of these dimensions seems suboptimal. One of the reasons for the success of CNNs is their invariance to small translations and scaling. Intuitively, small translations (corresponding to the pitch of voice) or scaling (corresponding to speaking slowly or quickly) should not change the class assigned to a window of speech. We hypothesise that classification of windows of speech with CNNs can be done more effectively with architectures similar to ones used in object recognition.

\begin{SCfigure}
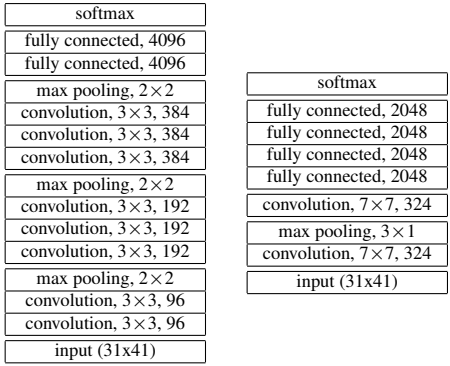

	\scriptsize
	\begin{tabular}{ | c | }
  	\hline
	softmax \\ \hline \hline
	fully connected, 4096 \\ \hline	
	fully connected, 4096 \\ \hline \hline

	max pooling, 2$\times$2 \\ \hline 
	convolution, 3$\times$3, 384 \\ \hline
	convolution, 3$\times$3, 384 \\ \hline
	convolution, 3$\times$3, 384 \\ \hline \hline

	max pooling, 2$\times$2 \\ \hline 
	convolution, 3$\times$3, 192 \\ \hline
	convolution, 3$\times$3, 192 \\ \hline
	convolution, 3$\times$3, 192 \\ \hline \hline

	max pooling, 2$\times$2 \\ \hline 
	convolution, 3$\times$3, 96 \\ \hline
	convolution, 3$\times$3, 96 \\	\hline \hline

	input (31x41) \\ \hline	
	\end{tabular}
	\qquad
	\begin{tabular}{ | c | }
  	\hline
	softmax \\ \hline \hline
	fully connected, 2048 \\ \hline	
	fully connected, 2048 \\ \hline
	fully connected, 2048 \\ \hline	
	fully connected, 2048 \\ \hline \hline

	convolution, 7$\times$7, 324 \\ \hline \hline

	max pooling, 3$\times$1 \\ \hline 
	convolution, 7$\times$7, 324 \\	\hline \hline

	input (31x41) \\ \hline	
	\end{tabular}
	\caption{Left panel: configuration of the vision-style convolutional network used in our experiments. We used the stride of two pixels for max-pooling layers and one pixel for convolutional layers. We used no zero-padding in the first two convolutional layers and one pixel zero-padding for all remaining convolutional layers. Right panel: configuration of the CNN very closely resembling the one in the work of \citet{LVCSR}, adjusted to match the number of parameters in our network.}
	\label{fig:convnet_configuration}
\end{SCfigure}

Looking at this problem through the lens of computer vision, we use a convolutional network architecture inspired by the work of \citet{very_deep_vision}. We only use small convolutional filters of size 3$\times$3, non-overlapping 2$\times$2 pooling regions and our network also has more layers than networks previously considered for the purpose of speech recognition. The same architecture is shared between both baseline and student networks (described in detail in \autoref{fig:convnet_configuration}, contrasted to a widely applied architecture proposed by \citet{LVCSR}).

\section{Combining bidirectional LSTMs with vision-style CNNs}

Both LSTMs and CNNs are powerful models, but the mechanisms that guide their learning are quite different. That creates an opportunity to combine their predictions, implicitly averaging their inductive biases. A classic way to perform this is ensembling, that is, to mix posterior predictions of the two models in the following manner:
\begin{equation*}
p(y | \mathbf{x}_i) = \gamma p_{\mathrm{LSTM}}(y | \mathbf{x}_i) + (1 - \gamma)p_{\mathrm{CNN}}(y | \mathbf{x}_i),
\label{eq:objective_ensemble}
\end{equation*}
where $\gamma \in [0, 1]$. The notation $p_{\mathrm{LSTM}}(y | \mathbf{x}_i)$ and $p_{\mathrm{CNN}}(y | \mathbf{x}_i)$ denotes probabilities of class $y$ given a feature vector $\mathbf{x}_i$, respectively for the LSTM and the baseline CNN. It is interesting to combinte these two types of models because they seem to meet the conditions of \citet{ensembles}: ``a necessary and sufficient condition for an ensemble of classifiers to be more accurate than any of its individual members is if the classifiers are accurate and diverse". Although ensembling is known to be very successful, it comes at the cost of executing all models at test time.

We propose an alternative which is to use model compression. Because capacity of the two models is similar we call it ``model blending'' rather than ``model compression''. To combine the inductive biases of both the LSTM and the CNN, we can use a training objective that combines the loss function on the hard labels from the training data with a loss function which penalises deviation from predictions of the LSTM teacher. That is, we optimise
\begin{equation}
L(\lambda) = \lambda \left[ -\sum_{i}\sum_{c} p_\mathrm{LSTM}(c | \mathbf{x}_i)\log q_\mathrm{CNN}(c | \mathbf{x}_i) \right] + (1 - \lambda) \left[-\sum_{i} \log q_\mathrm{CNN}({y_i} | \mathbf{x}_i) \right],
\label{eq:objective}
\end{equation}
where $p_\mathrm{LSTM}(c | \mathbf{x}_i)$ is the probability of class $c$ for training example $\mathbf{x}_i$ estimated by the teacher, $q_\mathrm{CNN}(c | \mathbf{x}_i)$ is the probability of class $c$ assigned to training example $\mathbf{x}_i$ by the student and $y_i$ is the correct class for $\mathbf{x}_i$. The coefficient $\lambda \in [0, 1]$ controls the weight of the errors on soft and hard labels in the objective. When $\lambda = 0$ the network is only learning using the hard labels, ignoring the teacher, while $\lambda = 1$ means that the networks is only learning from the soft labels provided by the teacher, ignoring the hard labels. When $\lambda \in (0, 1)$  optimising the objective in \autoref{eq:objective} yields a form of hybrid model which is learning using the guidance of the teacher, although not depending on it alone. With a symmetric objective, we could train an LSTM using the guidance of the CNN. Instead, we blend into the CNN for efficency at test time.

We motivate the choice of working directly with probabilities instead of logits (cf. \autoref{sec:background}) in two ways.
First, it is more direct to interpret retaining a subset of predictions of a network when considering probabilities (cf. \autoref{eq:subset}).
We can simply look at the fraction of probability mass a subset of outputs covers. This will be necessary in our work (see \autoref{sec:experiments}). Secondly, when using both soft and hard targets, it is easier to find an appropriate $\lambda$ and learning rate, when the two objectives we are weighting together are of similar magnitudes and optimisation landscapes.

\section{Experiments}
\label{sec:experiments}

\begin{figure}
      \centering
      \begin{minipage}{0.48\textwidth}
	\centering
	\begin{tabular}{cc}
  	\includegraphics[scale=0.23, trim=60 0 40 15]{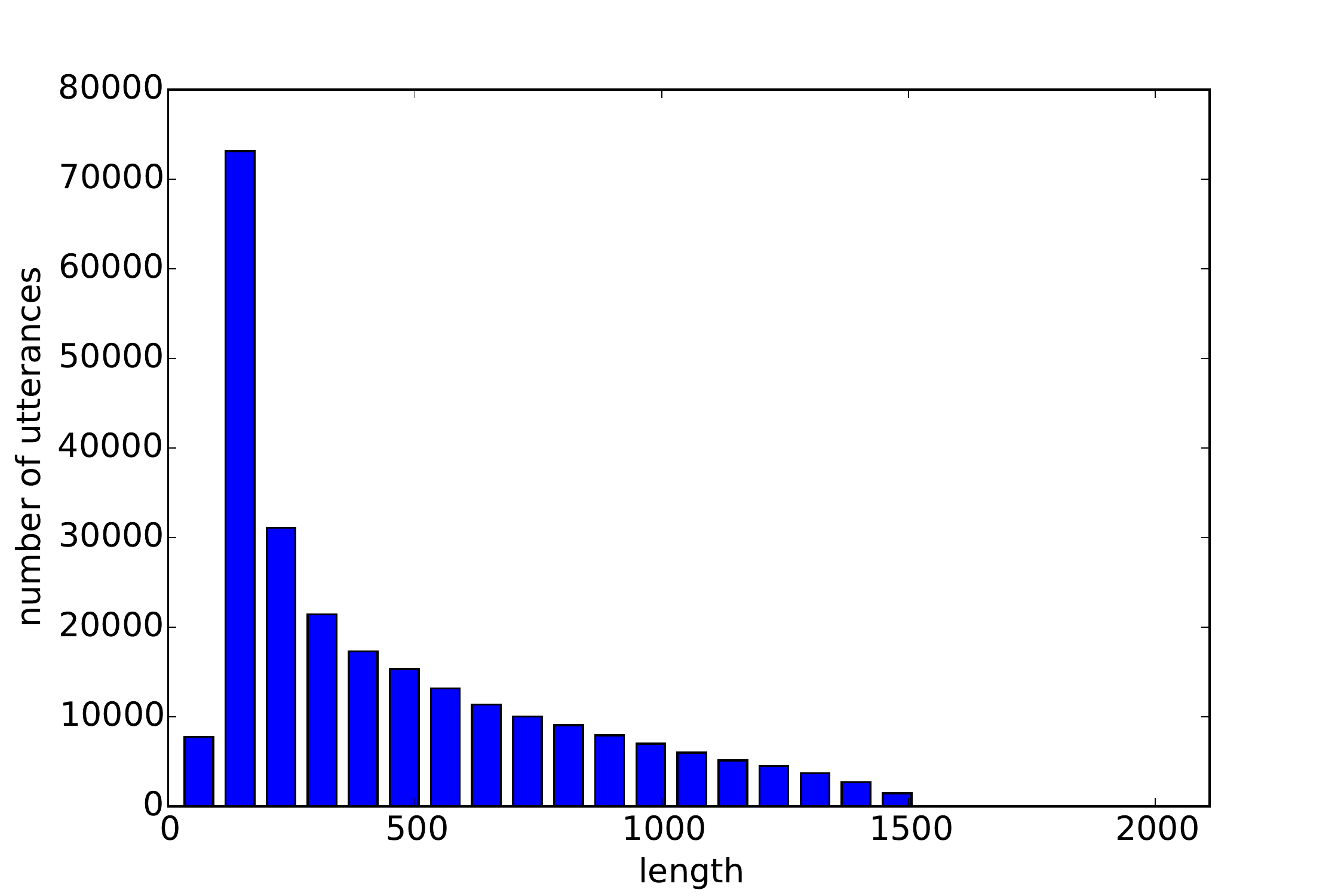} & \includegraphics[scale=0.23, trim=35 0 0 15]{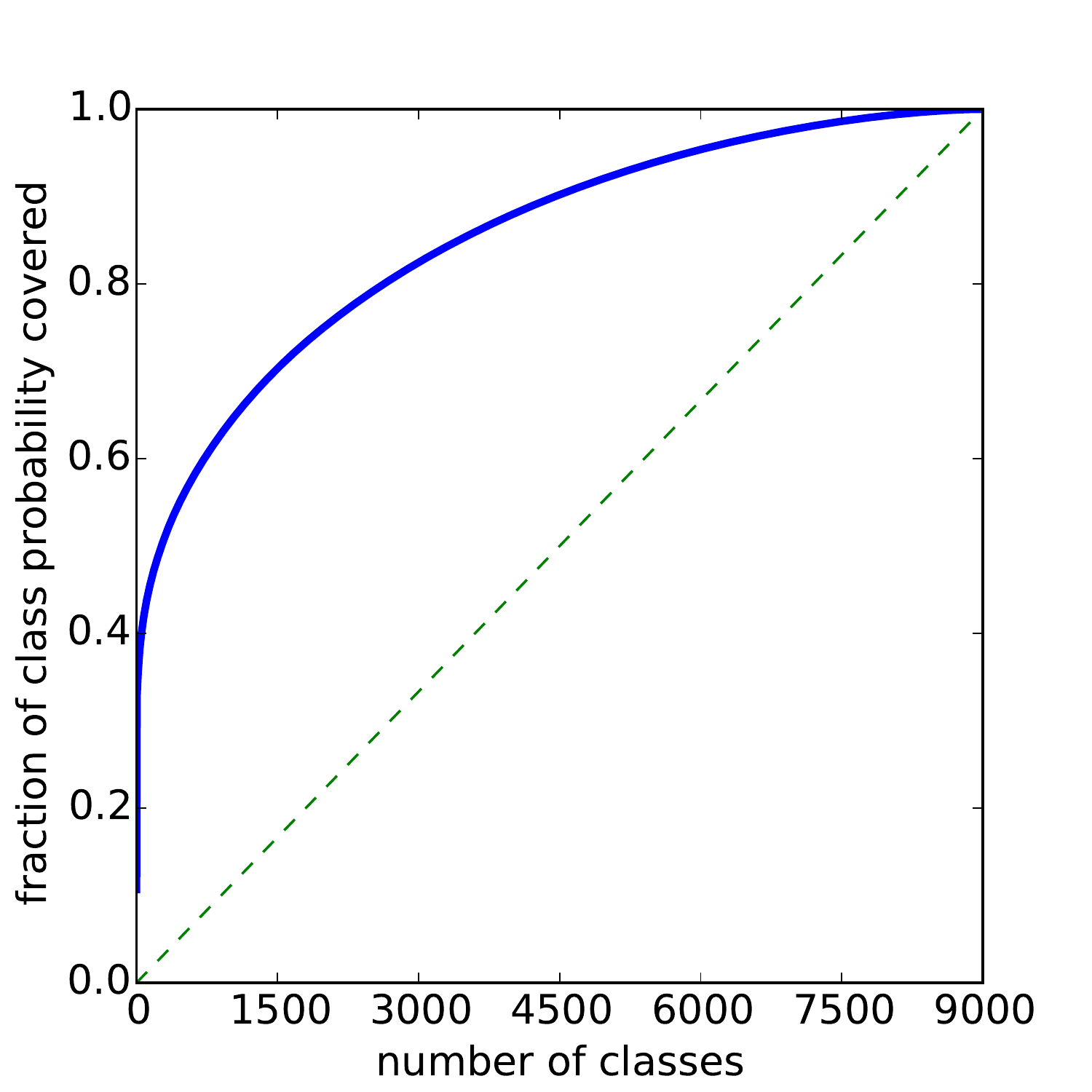}
	\end{tabular}
    \end{minipage}	
\begin{minipage}{0.48\textwidth}
	\centering
	\hspace{1cm}
	\includegraphics[scale=0.23, trim=50 0 130 15]{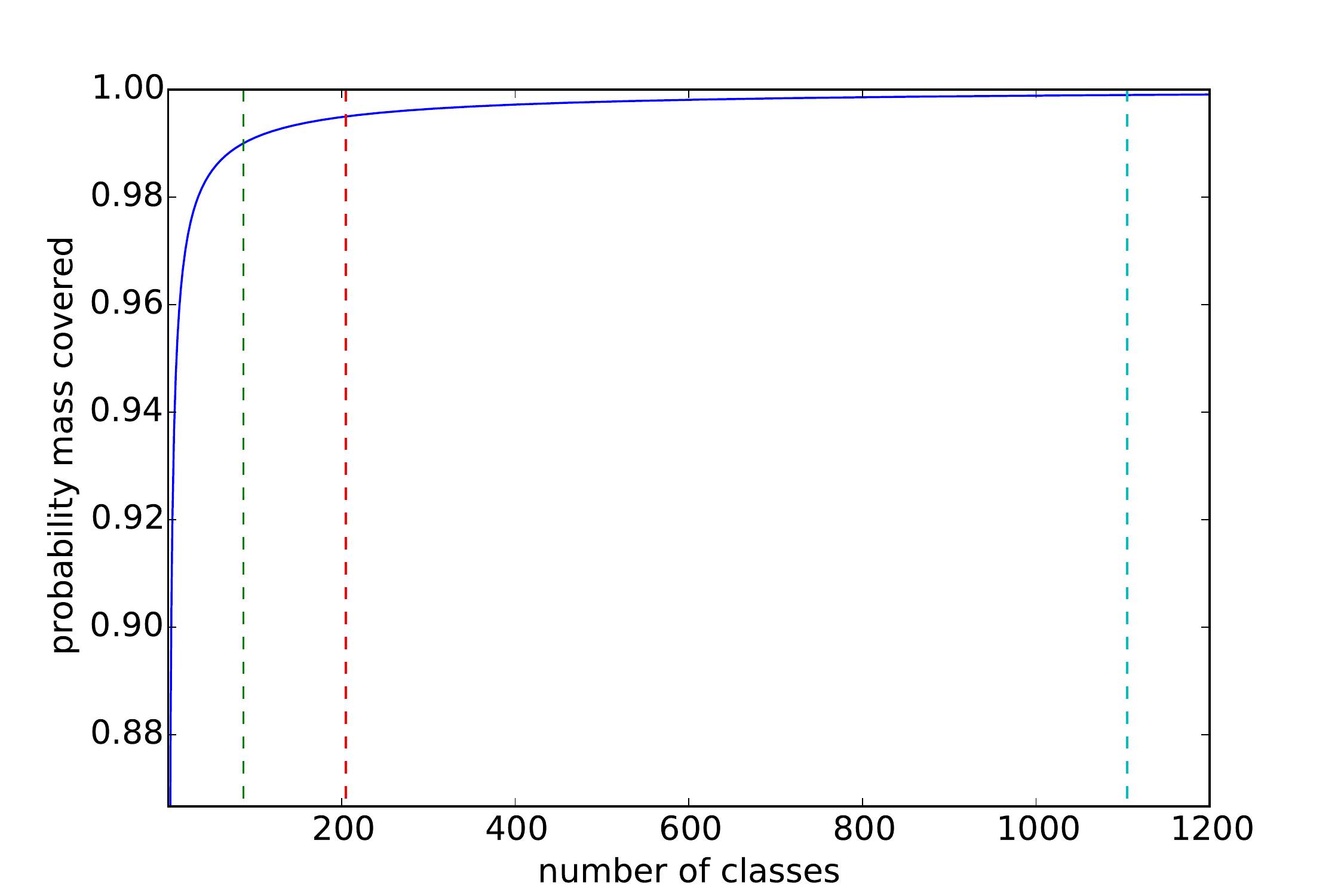}
    \end{minipage}
	\caption{Left: distribution of the lengths (numbers of frames) of utterances in the training data. Center: fraction of class probability in the data covered as a function of the number of most likely classes included. Dashed line indicates what the curve would look like if the labels had a uniform distribution. The distribution is very non-uniform. Right: average fraction of probability mass of the teacher LSTM predictions covered as a function of the number of most probable classes retained. Dashed lines indicate the number of classes necessary to cover 99\%, 99.5\% and 99.9\% of probability mass. Clearly, only very few top classes in the total of 9000 are necessary to cover a large majority of probability mass.}
	\label{fig:data}	\label{fig:percentage_covered}\vspace{-1ex}
\end{figure}

\begin{figure*}[h!]
	\centering
	\includegraphics[trim=25mm 68mm 21mm 70mm, clip, scale=0.87]{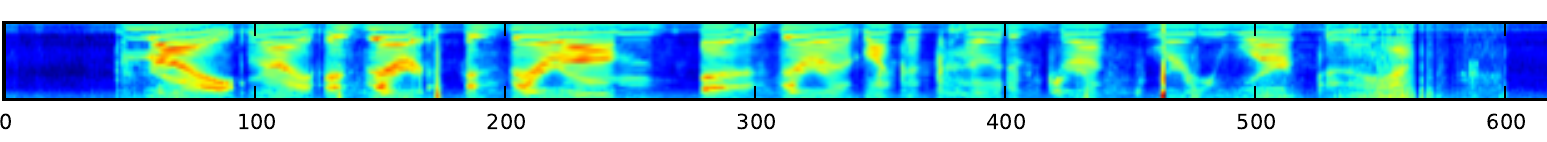}	
	\caption{Example utterance from Switchboard. The $x$-axis indicates time and the $y$-axis frequency.}
	\label{fig:example_utterance}\vspace{-1ex}
\end{figure*}

In our experiments we use the Switchboard data set \citep{switchboard}, which consists of 309 hours of transcribed speech. We used 308 hours as training set and kept one hour as a validation set. The data set is segmented into 248k utterances, i.e. continuous pieces of speech beginning and ending with a clear pause. Each utterance consists of a number of frames, i.e. 25 ms intervals of speech, with a constant shift of 10 ms. For every frame, the extracted features are 31-channel Mel-filterbank parameters passed through a 10-th root nonlinearity. Features for one utterance are visualised in \autoref{fig:example_utterance}. To form our training and validation sets we extract windows of 41 frames, that is, the frame whose label we want to predict, 20 frames before it and 20 frames after it. As shown in \autoref{fig:data}, distribution of the lengths of utterances is highly non-uniform, therefore to keep the sampling unbiased, we sample training examples by first sampling an utterance proportionally to its length and then sampling a window within that utterance uniformly. To form the validation set we simply extract all possible windows. In both cases, we pad each utterance with zeros at the beginning and at the end so that every frame in each utterance can be drawn as a middle frame. Every frame in the training and validation set has a label. The 9000 output classes represent tied tri-phone states that are generated by an HMM/GMM system \citep{hmm_state_tying}. Forced alignment is used to map each input frame to one output state using a baseline DNN model. The distribution of classes in the training data is visualised in \autoref{fig:data}. We call the frame classification error on the validation set frame error rate (FER). 

The test set is a part of the standard Switchboard benchmark (Hub5'00 SW). It was sampled from the same distribution as the training set and consists of 1831 utterances.
There are no frame-level labels in the test set and the final evaluation is based on the ability to predict words in the test utterances. To obtain the words predicted by the model, frame label posteriors generated from the neural network are first divided by their prior probabilities then passed to a finite state transducer-based decoder to be combined with 3-gram language model probabilities to generate the most probable word sequences. Hypothesized word sequences are aligned to the human reference transcription to count the number of word insertions ($I$), deletions ($D$), and substitutions ($S$). Word error rate (WER) is defined as $\mathrm{WER} = \frac{S+D+I}{N} \times 100$, where $N$ is the total number of words in the reference transcription.

Since the teacher model is very slow at prediction time, it is impractical to run a large number of experiments if the teacher must repeatedly be executed to train each student. Unfortunately, running the teacher once and saving its predictions to disk is also problematic --- because the output space is large (9000 classes), storing the soft labels for all classes would require a large amount of space ($\approx$ 3.6 TB). Moreover, to sample each minibatch in an unbiased manner, we would need constant random access to disk, which again would make training very slow. To deal with that problem we save predictions only for the small subset of classes with the highest predicted probabilities. To determine whether this is a viable solution, we checked what percentage of the total probability mass, averaged over the examples in the training set, is covered by the $C$ most likely classes according to the teacher model. We denote that set $\mathrm{TOP}_C(\mathbf{x})$. That is, we compute
\begin{equation}
\mathrm{M}(C) = \frac{1}{|\{\mathbf{x}_i\}|}\sum_{\mathbf{x}_i}\sum_{y \in \mathrm{TOP}_C(\mathbf{x}_i)} p(y | \mathbf{x}_i),
\label{eq:subset}
\end{equation}
where $p(y | \mathbf{x})$ denotes the posterior probability of class $y$ given a feature vector $\mathbf{x}_i$. This relationship for one of our LSTM models is shown in \autoref{fig:percentage_covered}. We found that, with very few exceptions, posteriors over classes are concentrated on very few values. Therefore, we decided to continue our experiments retaining top classes covering not more than 90 classes for each training example, cutting off after covering 99\% of the probability mass. 
This allows us to store soft labels for the entire data set in the RAM, making unbiased sampling of training data efficient.

\subsection{Baseline networks}

We used Lasagne, which is based on Theano \citep{theano_1}, for implementing CNNs. We used the architecture described in \autoref{fig:convnet_configuration}. For training of the CNN baseline we used mini-batches of size 256. Each epoch consisted of 2000 mini-batches. Hyper-parameters of the baseline networks were: initial learning rate ($1.7 \times 10^{-2}$), momentum coefficient (0.9, we used Nesterov's momentum) and a learning rate decay coefficient (0.7). Because the data set we used is very large (309 hours, 18 GB), the only form of regularisation we used was early stopping in the following form. After every epoch we measured the loss on the validation set. If the loss did not improve for five epochs, we multiplied the learning rate by a learning rate decay coefficient. We stopped the training after the learning rate was smaller than $5 \times 10^{-5}$. It took about 200 epochs to finish. We repeated training with three different random seeds.
For comparison, we also trained a CNN very similar to the one proposed by \citet{LVCSR}, adjusted to match the number of parameters in our network. We used the same training procedure and hyperparameters.

The bidirectional LSTM teacher networks we used are very similar to the one in the work of \cite{baseline_LSTM}. We trained three models, all with four hidden layers, two with 512 hidden units for each direction and one with 800 hidden units for each direction. The training starts with the learning rate equal to 0.05 for one of the smaller models and 0.08 for the other ones. We used the standard momentum with the coefficient of 0.9. After three epochs of no improvement of frame error rate on the validation set, the learning rate was multiplied by $\frac{2}{3}$ and training process was rolled back to the last epoch which improved validation error. Training stopped when learning rate was smaller than $10^{-5}$. It took about 75 epochs (2000 minibatches of 256 samples) to finish the training.

The results for these models are shown in \autoref{tab:results}. Our vision-style CNN achieved 14.1 WER (averaged over three random seeds), the larger LSTM achieved 14.4 WER, and a CNN of an architecture proposed by \citet{LVCSR} achieved 15.5 WER. Interestingly, although our LSTM teachers outperform our vision-style CNN trained with hard labels in terms of FER, the results in WER, which is the metric of primary interest, are the opposite. This discrepancy between FER and WER has been observed in the speech community before, for example by \citet{google_LSTM} for LSTMs and DNNs. FER and WER are not always perfectly correlated because FER is conditioned on a model that generated the frame alignment in the first place (which might not be correct for all the cases). FER also penalizes misclassifications of boundary frames which might not be of importance as long as the correct target state is recognized. On the other hand, WER is calculated taking into account information about neighbouring frames (i.e. smoothness) as well as external knowledge (e.g. a language model) which corrects many of the misclassifications made locally. 

\begin{table*}[htb!]
\footnotesize
\caption{FER and WER for our models. The numbers for the vision-style CNNs and LSTM$\to$CNN blending are averages over three random seeds. The numbers for smaller LSTM are given for a better of the two on FER, the other one achieved 34.71\% FER and the same WER.}
\label{tab:results}
\centering
\begin{tabular}{l|c|c|c|c|}
\cline{2-5}
& \textbf{FER} & \textbf{WER} & model size & execution time \\ \hline 
\multicolumn{1}{|l|}{\citet{LVCSR}-style CNN } & 37.93\% & 15.5 & $\approx$ 75M & $\times$ 0.75\\ \hline 
\multicolumn{1}{|l|}{vision-style CNN} & 35.51\% & 14.1 & $\approx$ 75M & $\times$ 1.0 \\\hline 
\multicolumn{1}{|l|}{smaller LSTM} & 34.27\% & 14.8 &$\approx$ 30M & $\times$ 3.3\\ \hline 
\multicolumn{1}{|l|}{bigger LSTM} & 34.15\% & 14.4 & $\approx$ 65M & $\times$ 5.8\\ \hline \hline
\multicolumn{1}{|l|}{LSTM + CNN ensemble ($\gamma = 0.5$)} & 32.4\% & 13.4 & $\approx$ 130M & $\times$ 6.8\\ \hline
\multicolumn{1}{|l|}{LSTM $\to$ CNN blending ($\lambda =0.75 $)} & 34.11\% & 13.83& $\approx$ 75M & $\times$ 1.0\\ \hline
\end{tabular}
\end{table*}

\subsection{Ensembles of networks}

\begin{figure}
    \centering
    \begin{minipage}{0.48\textwidth}
	\centering
	\includegraphics[scale=0.24, trim=0 10 0 20]{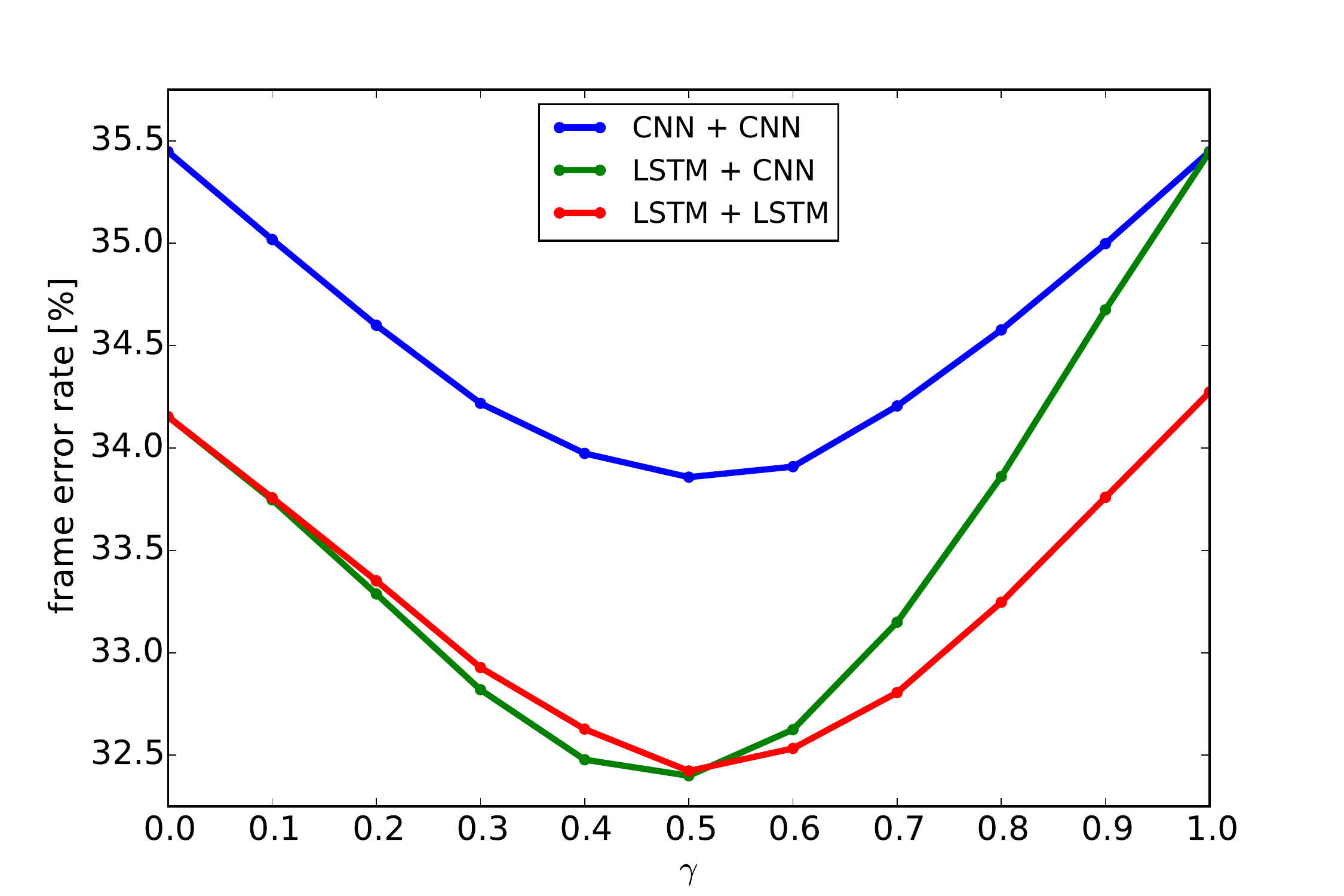}
	\caption{FER of ensembles as a function of $\gamma$.}
	\label{fig:fer_ensemble}
    \end{minipage}
	\begin{minipage}{.02\textwidth}		
		\hspace{1cm}
	\end{minipage}	
    \begin{minipage}{0.48\textwidth}
	\centering
	\includegraphics[scale=0.24, trim=0 10 0 20]{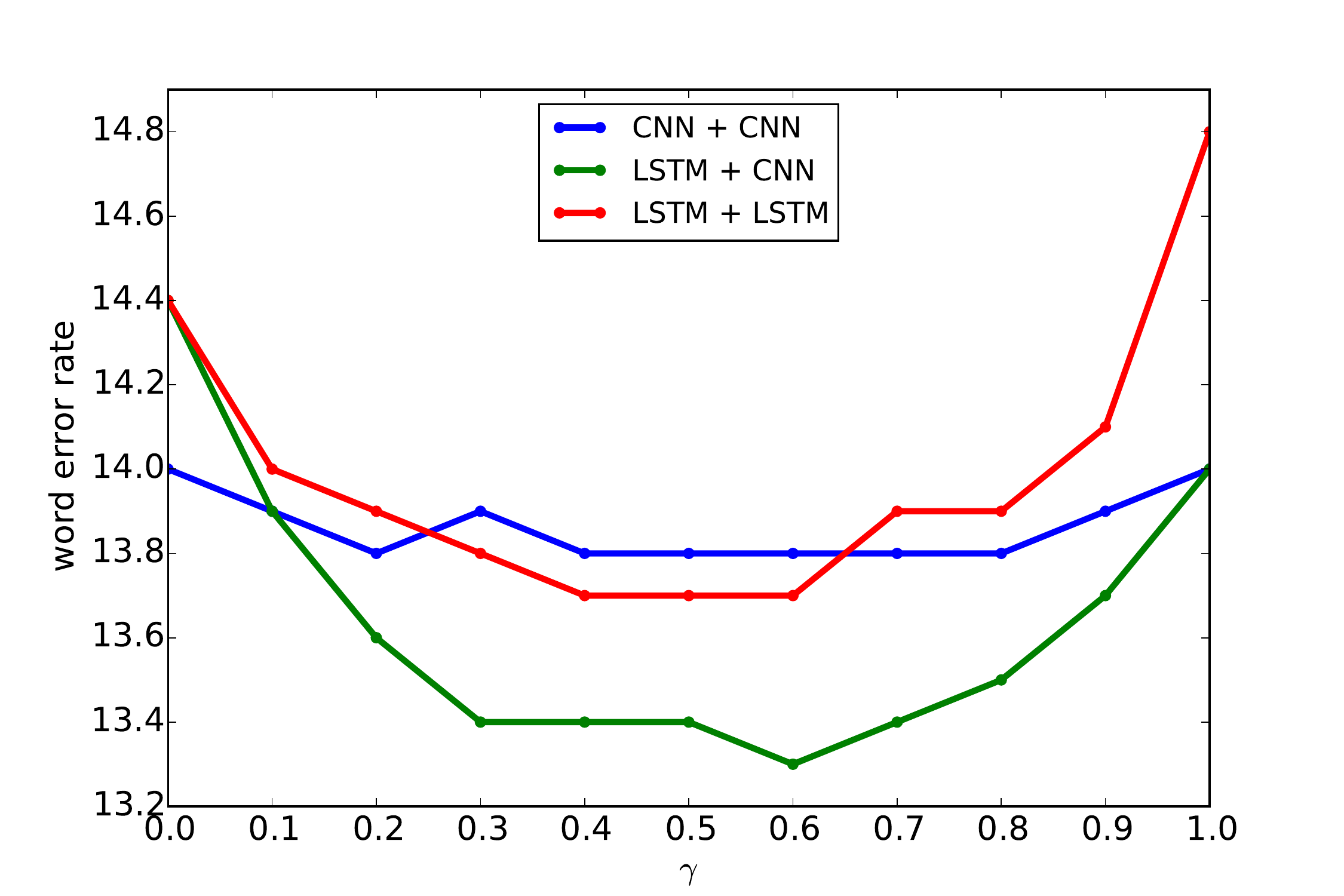}		
	\caption{WER of ensembles as a function of $\gamma$.}
	\label{fig:wer_ensemble}
    \end{minipage}
\end{figure}

The first approach we use to combine the two types of models is to create ensembles. The results in \autoref{fig:fer_ensemble} and \autoref{fig:wer_ensemble} indicate that for the problem we consider it is beneficial to combine neural networks from different families, which have different inductive biases. Even though CNNs are much weaker in FER, combining an LSTM with a vision-style CNN achieves the same FER as an ensemble of two LSTMs (both of which are more accurate than the CNN), and actually yields better WER than ensembles of two CNNs or two (superior) LSTMs. Interestingly, ensembling two CNNs yields almost no benefit in WER. To complete the picture we tried ensembles with more than two models. An ensemble of three LSTMs achieved 31.98\% FER and 13.7 WER, an ensemble of three CNNs achieved 33.31\% FER and 13.9 WER, thus, in both cases, yielding very little gain over ensembles of two models of these types. On the other hand adding a CNN to the ensemble of two LSTMs yielded 31.57\% FER and 13.2 WER. The benefits of adding more models to the ensemble appear to be negligible if the ensemble contained at least one model of each type already. Although the ensembles we trained are very effective in terms of WER, it comes at the cost of a large increase of computation at test time compared to the baseline CNN. Because of the cost of the LSTMs, our best two-model ensemble is about 7 times slower than our vision-style CNN (cf. \autoref{tab:results}).

We also compared the errors made by CNNs and LSTMs to see if the models are qualitatively different. We observe that a CNN tends to make similar errors as other CNNs, an LSTM tends to make similar errors as other LSTMs, but CNNs and LSTMs tend to make errors that are less similar to each other than the CNN-CNN and LSTM-LSTM comparisons. Overall, we conclude that the inductive biases of LSTM and CNN are complementary.

\subsection{Networks trained with model blending via model compression}

\begin{table*}
\footnotesize
\caption{Average fraction of probability mass covered and average number of classes retained after truncating predictions of an ensemble of two LSTMs to fit them in the memory.}
\label{tab:fraction_left}
\centering
\begin{tabular}{l|c|c|c|c|c|}
\hline
\multicolumn{1}{|l|}{\textbf{maximum number of classes retained}} & \textbf{1} & \textbf{3} & \textbf{10} & \textbf{30} & \textbf{90}\\ \hline 
\multicolumn{1}{|l|}{average fraction of probability mass covered} & 73.20\% & 91.19\% & 96.68\% & 98.38\% & 99.03\% \\ \hline
\multicolumn{1}{|l|}{average number of classes retained} & 1.0 & 2.89 & 6.79 & 13.33 & 23.96 \\ \hline 
\end{tabular}
\end{table*}

The next question we tackle is whether it is possible to achieve an effect similar to creating an ensemble without having to execute all models at prediction time. We attack this with model blending via model compression. To do this, we took predictions of the two best performing LSTMs in \autoref{tab:results} and averaged their predictions to form a teacher model. Such a teacher model achieves 32.4\% FER and 13.4 WER. As we mentioned earlier, it is infeasible to use all predictions during training. Therefore we only store a subset of classes predicted by the teacher for each frame. \autoref{tab:fraction_left} shows what fraction of probability mass is covered when storing different maximum number of predictions ($C$). It is particularly interesting to understand how many predictions of the teacher model are sufficient to achieve good performance to test the \emph{dark knowledge} hypothesis \citep{distilling_knowledge} which states that information about classes predicted with low probability is important to the success of model compression. We use $C \in \{90, 30, 10, 3, 1\}$. We also vary the parameter $\lambda$ which controls how much the student is learning from the teacher and how much it is learning from hard labels. The architecture and training procedure for the students is the same as for the baseline. For every combination of $C$ and $\lambda$ we report an average over three random seeds.

\begin{figure}
    \centering
    \begin{minipage}{0.48\textwidth}
	\centering
	\includegraphics[scale=0.24, trim=0 10 0 20]{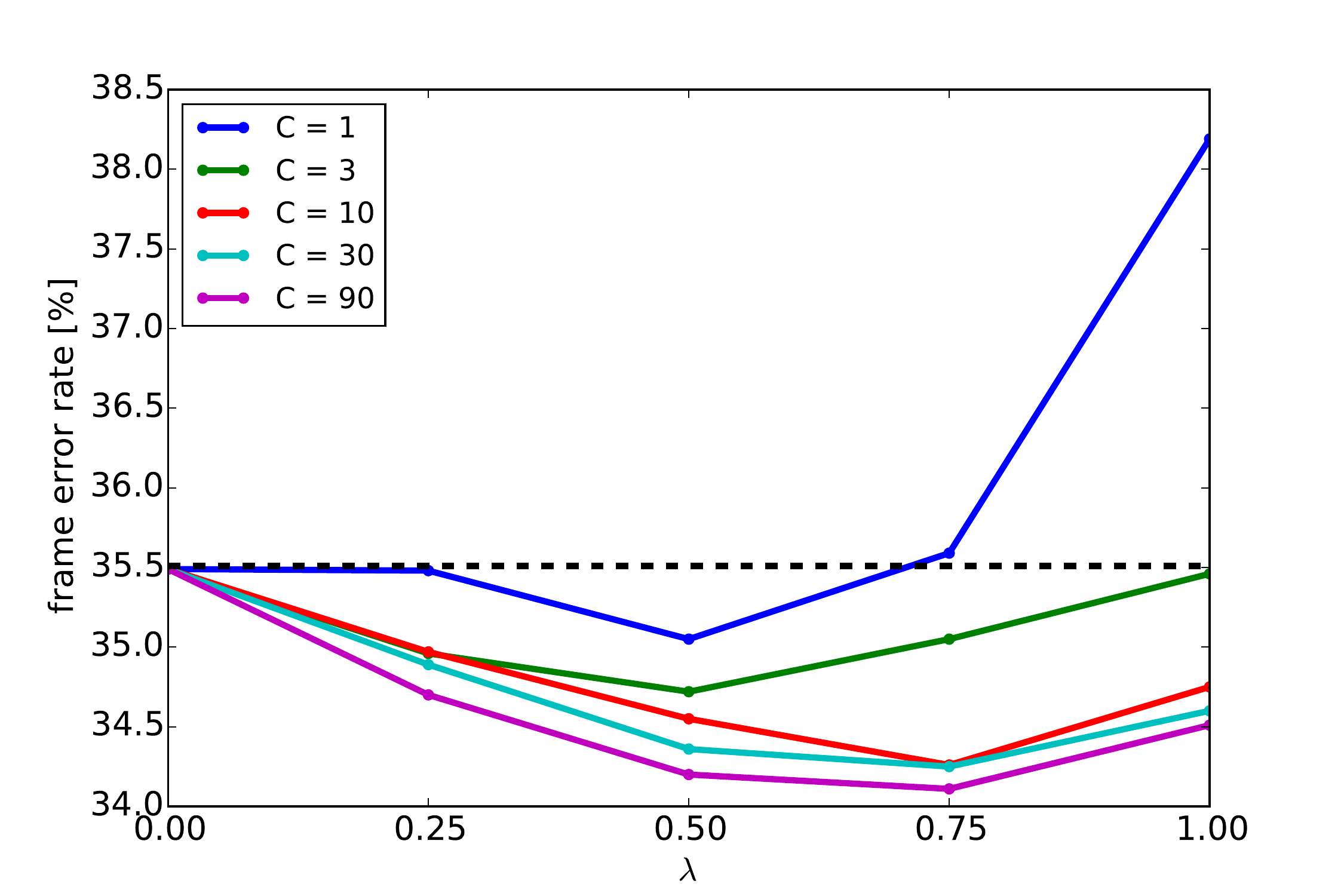}
	\caption{FER of the CNN student as a function of $\lambda$ as $C$, the maximum of number of teacher outputs retained, varies from 1 to 90. Dashed line indicates performance of a baseline CNN.}
	\label{fig:fer}
    \end{minipage}
	\begin{minipage}{.02\textwidth}		
		\hspace{1cm}
	\end{minipage}	
    \begin{minipage}{0.48\textwidth}
	\centering
	\includegraphics[scale=0.24, trim=0 10 0 20]{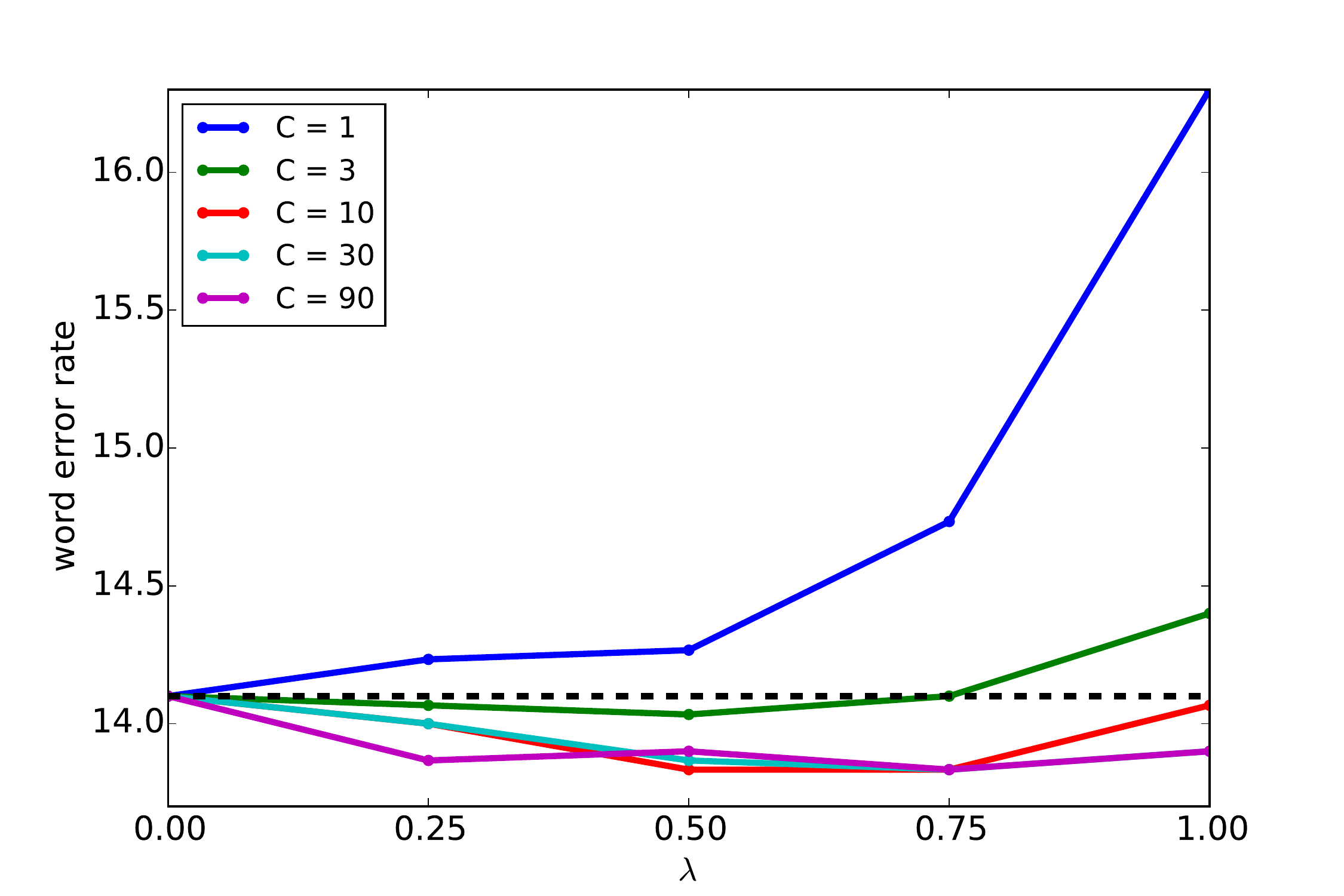}	
	\caption{WER of the CNN student as a function of $\lambda$ as $C$, the maximum of number of teacher outputs retained, varies from 1 to 90. Dashed line indicates performance of a baseline CNN.}
	\label{fig:wer}
    \end{minipage}
\end{figure}

The results are shown in \autoref{fig:fer} and \autoref{fig:wer}. The best model achieved lower FER (34.11\%) and lower WER (13.83) than any of the individual models. Furthermore, the blended model has fewer parameters and is 6.8 times faster at test time than the ensemble of the LSTM and the CNN. For all numbers of teacher predictions retained ($C$) the best performance was achieved for $\lambda \in \{0.25, 0.5, 0.75\}$. That highlights the importance of blending the knowledge extracted from the teacher model with learning from hard labels within the architecture of the student.

Our experiments show that, at least for this task, it is not critical to use the teacher predictions for all classes. Just the 30 most likely predictions ($\frac{1}{300}$ of all classes!) is enough. Bringing the number up to 90 classes did not improve the student WER performance.
However, performance deteriorates dramatically when too few predictions are used, suggesting that some dark knowledge is needed.

Finally, we experimented with using a CNN as a teacher for a CNN of the same architecture, i.e. we took predictions of a baseline vision-style CNN and used its predictions to train another CNN of the same architecture (with $\lambda=0.5$ and $C=90$). Such a student achieves on average (over 3 random trials) 34.61\% FER and 14.1 WER, which is, as we expected, worse than a student of the LSTMs since the two models are more similar, but still significantly better than the baseline in terms of FER. These results are consistent with the results for ensembles (cf. \autoref{fig:fer_ensemble} and \autoref{fig:wer_ensemble}). Clearly, blending dissimilar models like CNNs and LSTMs is stronger.

\section{Related work}
A few papers have applied model compression in speech recognition settings. The most similar is by \citet{transferring_knowledge} who compressed LSTMs into small DNNs without convolutional layers. Using the soft labels from an LSTM, they were able to show an improvement in WER over the baseline trained with hard labels. The main difference between this work and ours is that their students are non-convolutional and tiny. While this allows for a decent improvement over the baseline, since the student network is much smaller, its performance is still much weaker than performance of a single model of the same type as the teacher. Hence, this work addresses a different question than we do, i.e. whether a network without recurrent structure can perform as well or better as an LSTM when using soft labels provided by the LSTM. Model compression was also successfully applied to speech recognition by \citet{small_size_dnn} who used DNNs without convolutional layers both as a teacher and a student. The architecture of the two networks was the same except that the student had less hidden units in each layer. Finally, work in the opposite direction was done by \citet{dark_knowledge_transfer} and \citet{knowledge_transfer}. They demonstrated that when using a small data set for which an LSTM is overfitting, a deep non-convolutional network can provide useful guidance for the LSTM. It can come either in the form of pre-training the LSTM with soft labels from a DNN or training the LSTM optimising a loss mixing hard labels with soft labels from a DNN. We are not aware of previous work on model compression in the setting, in which the student and the teacher are of similar capacity.

\section{Discussion}

The main contribution of this paper is introducing the use of model compression in an unexplored setting where both the teacher and student architectures are powerful ones, yet with different inductive biases. Thus, rather than calling it model compression we use the term ``model blending''. We showed that the LSTM and the CNN  learn different kinds of knowledge from the data which can be leveraged through simple ensembling or model blending via model compression. We provided experimental evidence that CNNs of appropriate vision-style architecture have the necessary capacity to learn accurate predictors on large speech data sets and gave a simple, practical recipe for improving the performance of CNN-based speech recognition models even further at no cost during test time. We hypothesise that the very recent advances in training even deeper convolutional networks for computer vision \citep{highway, deep_residual} will yield improved performance in speech recognition and would further improve our results. Finally, by using a CNN to teach a CNN, we have shown a very easy way of improving a neural network without training networks of more than one architecture or even forming ensembles.

\subsubsection*{Acknowledgments}
We thank Stanis{\l}aw Jastrz\k{e}bski for suggesting the name of the paper. We also thank Steve Renals and Pawe\l{} \'Swi\k{e}toja\'nski for insightful comments.

\bibliography{bibliography}
\bibliographystyle{iclr2016_workshop}

\newpage
\section*{Supplementary material}
\section*{Details of the LSTM}

Given a sequence of input vectors $\inseq = (x_1,\ldots,x_T)$, an RNN computes the hidden vector sequence $\seq{h} = (h_1,\ldots,h_T)$ by iterating the following from $t=1$ to $T$:
\begin{align*}
h_t &= \hiddenfn\left(\ihwts x_t + \hhwts h_{t-1} + \hbias \right).
\end{align*}

The $W$ terms denote weight matrices (\eg $\ihwts$ is the input-hidden weight matrix), the $b$ terms denote bias vectors (\eg $\hbias$ is hidden bias vector) and $\hiddenfn$ is the hidden layer function.

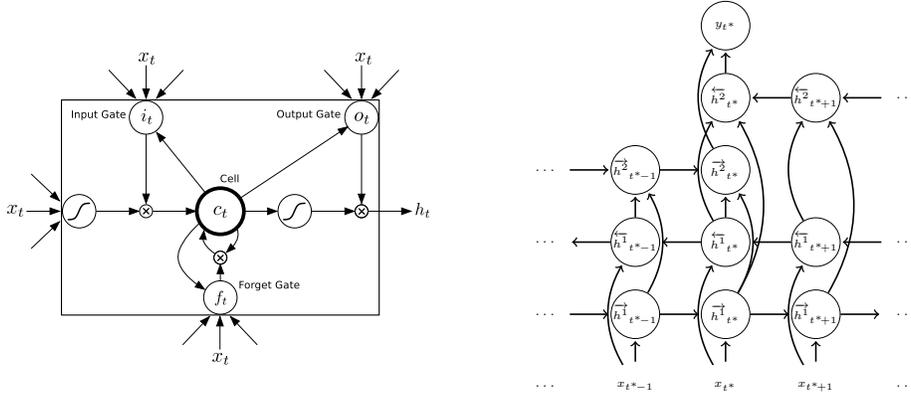
\begin{figure}[h!]
\centering
	\begin{tabular}{ccc}
  	\includegraphics[scale=0.35, trim=0 -50 0 0]{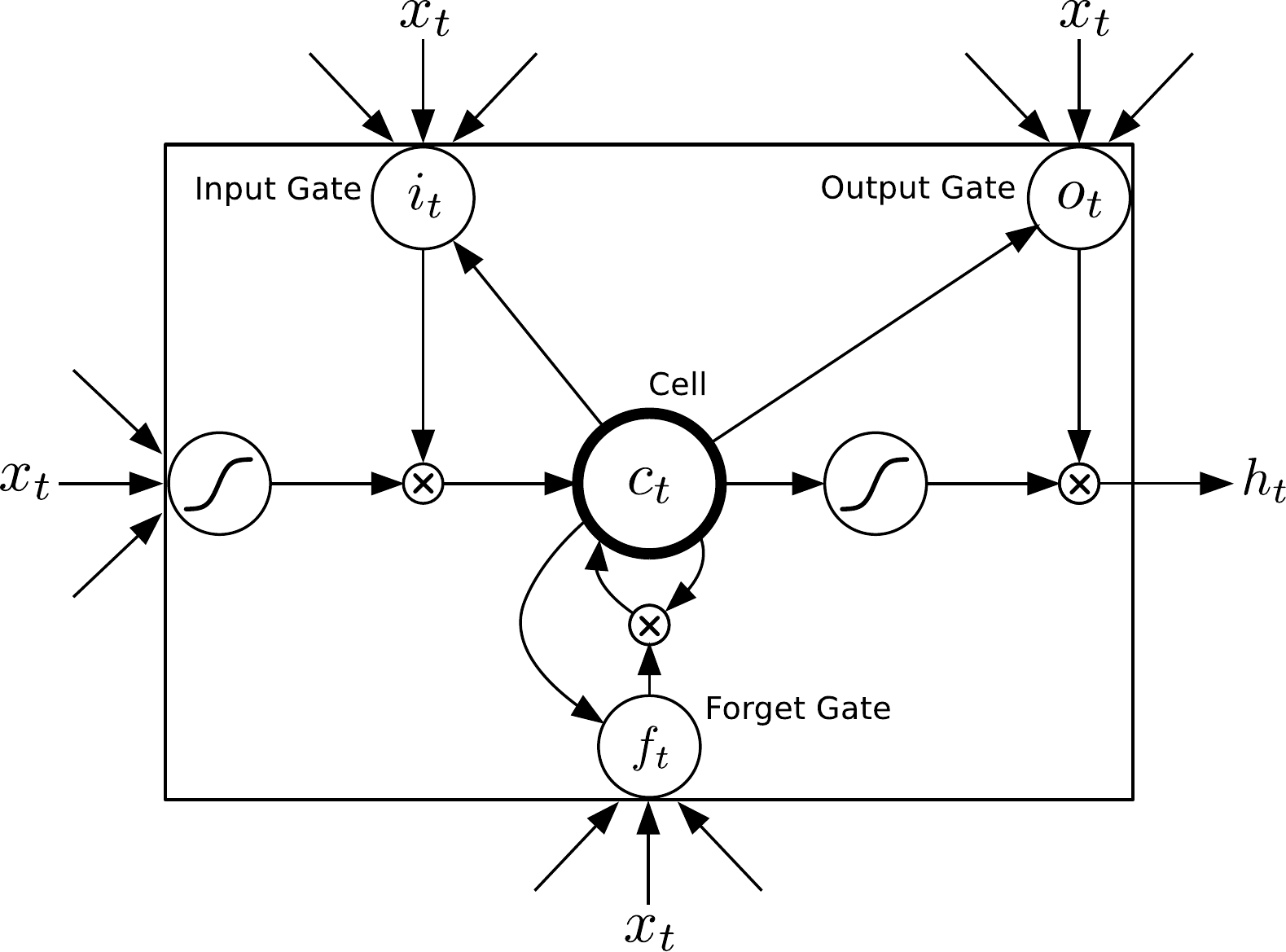} & \hspace{0.1cm} & \tiny
\def\layersep{1.2cm}
\def\scaling{1.5cm}
\def\arrowcolor{black}
\scalebox{0.8}{
\begin{tikzpicture}[shorten >=1pt,->, draw=black, node distance=\layersep]
    \tikzstyle{neuron}=[circle, draw=black, fill=white, minimum size=23pt, inner sep=0pt]
    \tikzstyle{invisible neuron}=[circle, draw=white, fill=white, minimum size=23pt, inner sep=0pt]
    \tikzstyle{annot}=[text width=4em, text centered]

    \foreach \name / \y in {0,...,4}
        \node[invisible neuron] (X-\name) at (\scaling*\y, 0) {};

    \foreach \name / \y in {1,...,3}
        \node[neuron] (F1-\name) at (\scaling*\y, \layersep) {};

    \foreach \name / \y in {0,4}
        \node[invisible neuron] (F1-\name) at (\scaling*\y, \layersep) {};

    \foreach \name / \y in {1,...,3}
        \node[neuron] (B1-\name) at (\scaling*\y, 2*\layersep) {};

    \foreach \name / \y in {0,4}
        \node[invisible neuron] (B1-\name) at (\scaling*\y, 2*\layersep) {};

    \foreach \name / \y in {1,...,2}
        \node[neuron] (F3-\name) at (\scaling*\y, 3*\layersep) {};

    \foreach \name / \y in {0}
        \node[invisible neuron] (F3-\name) at (\scaling*\y, 3*\layersep) {};

    \foreach \name / \y in {2,...,3}
        \node[neuron] (B3-\name) at (\scaling*\y, 4*\layersep) {};

    \foreach \name / \y in {4}
        \node[invisible neuron] (B3-\name) at (\scaling*\y, 4*\layersep) {};

    \foreach \name / \y in {2}
        \node[neuron] (Y-\name) at (\scaling*\y, 5*\layersep) {};

    \foreach \i in {1,...,3}
        \path (X-\i) edge [draw=\arrowcolor, line width=0.75pt] (F1-\i);

    \foreach \i in {1,...,3}
        \path (X-\i) edge [bend left, draw=\arrowcolor, line width=0.75pt] (B1-\i);

    \foreach \i in {1,...,2}
        \path (F1-\i) edge [bend right, draw=\arrowcolor, line width=0.75pt] (F3-\i);

    \foreach \i in {2,...,3}
        \path (F1-\i) edge [bend right, draw=\arrowcolor, line width=0.75pt] (B3-\i);

    \foreach \i in {1,...,2}
        \path (B1-\i) edge [draw=\arrowcolor, line width=0.75pt] (F3-\i);

    \foreach \i in {2,...,3}
        \path (B1-\i) edge [bend left, draw=\arrowcolor, line width=0.75pt] (B3-\i);

    \path (F1-0.east) edge [draw=\arrowcolor, line width=0.75pt] (F1-1.west);
    \path (F1-1.east) edge [draw=\arrowcolor, line width=0.75pt] (F1-2.west);
    \path (F1-2.east) edge [draw=\arrowcolor, line width=0.75pt] (F1-3.west);
    \path (F1-3.east) edge [draw=\arrowcolor, line width=0.75pt] (F1-4.west);

    \path (B1-4.west) edge [draw=\arrowcolor, line width=0.75pt] (B1-3.east);
    \path (B1-3.west) edge [draw=\arrowcolor, line width=0.75pt] (B1-2.east);
    \path (B1-2.west) edge [draw=\arrowcolor, line width=0.75pt] (B1-1.east);
    \path (B1-1.west) edge [draw=\arrowcolor, line width=0.75pt] (B1-0.east);

    \path (F3-0.east) edge [draw=\arrowcolor, line width=0.75pt] (F3-1.west);
    \path (F3-1.east) edge [draw=\arrowcolor, line width=0.75pt] (F3-2.west);

    \path (B3-3.west) edge [draw=\arrowcolor, line width=0.75pt] (B3-2.east);
    \path (B3-4.west) edge [draw=\arrowcolor, line width=0.75pt] (B3-3.east);

    \path (F3-2) edge [bend left, draw=\arrowcolor, line width=0.75pt] (Y-2);
    \path (B3-2) edge [draw=\arrowcolor, line width=0.75pt] (Y-2);

    \node[annot, left of=Y-2, node distance=0.0cm] {\scalebox{0.9}{$y_{t^*}$}};

    \node[annot, left of=X-1, node distance=0.0cm] {\scalebox{0.9}{$x_{t^*\!-1}$}};
    \node[annot, left of=X-2, node distance=0.0cm] {\scalebox{0.9}{$x_{t^*}$}};
    \node[annot, left of=X-3, node distance=0.0cm] {\scalebox{0.9}{$x_{t^*\!+1}$}};

    \node[annot, left of=F1-1, node distance=0.0cm] {\scalebox{0.9}{$\overrightarrow{{h}^1}_{t^*\!-1}$}};
    \node[annot, left of=F1-2, node distance=0.0cm] {\scalebox{0.9}{$\overrightarrow{{h}^1}_{t^*}$}};
    \node[annot, left of=F1-3, node distance=0.0cm] {\scalebox{0.9}{$\overrightarrow{{h}^1}_{t^*\!+1}$}};

    \node[annot, left of=B1-1, node distance=0.0cm] {\scalebox{0.9}{$\overleftarrow{{h}^1}_{t^*\!-1}$}};
    \node[annot, left of=B1-2, node distance=0.0cm] {\scalebox{0.9}{$\overleftarrow{{h}^1}_{t^*}$}};
    \node[annot, left of=B1-3, node distance=0.0cm] {\scalebox{0.9}{$\overleftarrow{{h}^1}_{t^*\!+1}$}};

    \node[annot, left of=F3-1, node distance=0.0cm] {\scalebox{0.9}{$\overrightarrow{{h}^2}_{t^*\!-1}$}};
    \node[annot, left of=F3-2, node distance=0.0cm] {\scalebox{0.9}{$\overrightarrow{{h}^2}_{t^*}$}};

    \node[annot, left of=B3-2, node distance=0.0cm] {\scalebox{0.9}{$\overleftarrow{{h}^2}_{t^*}$}};
    \node[annot, left of=B3-3, node distance=0.0cm] {\scalebox{0.9}{$\overleftarrow{{h}^2}_{t^*\!+1}$}};

    \node[annot, left of=X-0, node distance=0.0cm] {$\ldots$};
    \node[annot, left of=X-4, node distance=0.0cm] {$\ldots$};

    \node[annot, left of=F1-0, node distance=0.0cm] {$\ldots$};
    \node[annot, left of=F1-4, node distance=0.0cm] {$\ldots$};

    \node[annot, left of=B1-0, node distance=0.0cm] {$\ldots$};
    \node[annot, left of=B1-4, node distance=0.0cm] {$\ldots$};

    \node[annot, left of=F3-0, node distance=0.0cm] {$\ldots$};
    \node[annot, left of=B3-4, node distance=0.0cm] {$\ldots$};
\end{tikzpicture}
} \\
	\end{tabular}
	\caption{Left panel: the LSTM cell. Figure from \citet{bidirectional_speech}. Right panel: two-layer bidirectional RNN with one output.}
	\label{fig:lstm}
\end{figure}

While there are multiple possible choices for $\hiddenfn$, prior work \citep{sequence_labelling, bidirectional_speech, google_LSTM} has shown that the LSTM architecture, which uses purpose-built \emph{memory cells} to store information, is better at finding and exploiting longer context. The left panel of \autoref{fig:lstm} illustrates a single LSTM memory cell.
For the version of the LSTM cell used in this paper \citep{learning_lstm} $\hiddenfn$ is implemented by the following composite function:
\begin{align*}
\igate_t &= \sigma\left(\wtmat{x}{\igate} x_t + \wtmat{h}{\igate} h_{t-1} + \wtmat{\state}{\igate} \state_{t-1}  + b_\igate\right),\\
\fgate_t &= \sigma\left(\wtmat{x}{\fgate} x_t + \wtmat{h}{\fgate} h_{t-1} + \wtmat{\state}{\fgate} \state_{t-1} + b_\fgate \right),\\
\state_t &= \fgate_t \state_{t-1} + \igate_t \tanh \left(\wtmat{x}{\state} x_t + \wtmat{h}{\state} h_{t-1} + b_\state \right),\\
\ogate_t &= \sigma\left(\wtmat{x}{\ogate} x_t + \wtmat{h}{\ogate} h_{t-1} + \wtmat{\state}{\ogate} \state_{t} + b_\ogate\right),\\
h_t &= \ogate_t \tanh(\state_t),
\end{align*}
where $\sigma$ is the logistic sigmoid function, and $\igate$, $\fgate$, $\ogate$ and $\state$ are respectively the \emph{input gate}, \emph{forget gate}, \emph{output gate} and \emph{cell} activation vectors, all of which are the same size as the hidden vector $h$. The weight matrices from the cell to gate vectors (\eg $\wtmat{c}{\igate}$) are diagonal, so element $m$ in each gate vector only receives input from element $m$ of the cell vector.

One shortcoming of conventional RNNs is that they are only able to make use of previous context.
Bidirectional RNNs (BRNNs) \citep{brnn} exploit past and future context by processing the data in both directions with two separate hidden layers, which are then fed forwards to the same output layer. A BRNN computes the \emph{forward} hidden sequence $\seq{\hfor} = (\hfor_1,\ldots,\hfor_T)$ and the \emph{backward} hidden sequence $\seq{\hback} = (\hback_1,\ldots,\hback_T)$ by iterating from $t=1$ to $T$:
\begin{align*}
\hfor_t &= \hiddenfn\left(\wtmat{x}{\hfor} x_t + \wtmat{\hfor}{\hfor} \hfor_{t-1} + \bias{\hfor} \right),\\
\hback_t &= \hiddenfn\left(\wtmat{x}{\hback} x_t + \wtmat{\hback}{\hback} \hback_{t+1} + \bias{\hback}\right).
\end{align*}
Combining BRNNs with LSTMs gives the bidirectional LSTM, which can access the context in both directions.

Finally, deep RNNs can be created by stacking multiple RNN hidden layers on top of each other, with the output sequence of one layer forming the input sequence for the next. Assuming the same hidden layer function is used for all $N$ layers in the stack, the hidden vector sequences $\seq{h}^n$ are iteratively computed from $n=1$ to $N$ and $t=1$ to $T$:
\begin{align*}
h^n_t &= \hiddenfn\left(W_{h^{n-1}h^{n}} h^{n-1}_t + W_{h^{n}h^{n}} h^n_{t-1} + \hbias^n \right),
\end{align*}
where we define $\seq{h}^0 = \inseq$.

Deep bidirectional RNNs can be implemented by replacing each hidden sequence $\seq{h}^n$ with the forward and backward sequences $\overrightarrow{\seq{h}^n}$ and $\overleftarrow{\seq{h}^n}$, and ensuring that every hidden layer receives input from both the forward and backward layers at the level below. If bidirectional LSTMs are used for the hidden layers we get deep bidirectional LSTMs, the architecture we use as a teacher network in this paper.

In this work, following \cite{baseline_LSTM}, we only predict the label of the middle frame, hence the network output is computed as $y_{t^*} = \wtmat{h^N}{y} h^N_{t^*} + \obias$. This also implies that in the last hidden layer the forward sequence runs only from 1 to $t^*$ and the backward sequence runs only from $T$ to $t^*$. This is illustrated in the right panel of \autoref{fig:lstm}.
\end{document}